\documentclass{article}

\usepackage{arxiv}

\usepackage[utf8]{inputenc} 
\usepackage[T1]{fontenc}    
\usepackage{hyperref}       
\usepackage{url}            
\usepackage{booktabs}       
\usepackage{amsfonts}       
\usepackage{nicefrac}       
\usepackage{microtype}      
\usepackage{lipsum}
\usepackage{graphicx}
\usepackage{subfig}

\title{NeuroEvolution algorithms applied in the designing process of biohybrid actuators}

\author{
    Hugo Alcaraz-Herrera\\
    Unconventional Computing Laboratory, \\
    College of Arts, Technology and Environment, \\
    University of the West of England, \\
    Bristol, BS16 1QY, United Kingdom \\ \texttt{hugo.alcaraz@uwe.ac.uk}
\And
    Michail-Antisthenis Tsompanas  \\
    Unconventional Computing Laboratory \& \\
    School of Computing \& Creative Technologies,\\
    College of Arts, Technology and Environment, \\
    University of the West of England,\\ 
    Bristol, BS16 1QY, United Kingdom \\
    \texttt{antisthenis.tsompanas@uwe.ac.uk}
\And
       Andrew Adamatzky \\
       Unconventional Computing Laboratory,\\
    College of Arts, Technology and Environment, \\
    University of the West of England,\\ 
    Bristol, BS16 1QY, United Kingdom \\
\And
       Igor Balaz\\
       Laboratory for Meteorology, Physics and Biophysics,\\ Faculty of Agriculture, \\
       University of Novi Sad, \\ Trg Dositeja Obradovica 8, 21000, Novi Sad, Serbia
}

\begin{document}
\maketitle

\begin{abstract}
Soft robots diverge from traditional rigid robotics, offering unique advantages in adaptability, safety, and human-robot interaction. In some cases, soft robots can be powered by biohybrid actuators and the design process of these systems is far from straightforward. We analyse here two algorithms that may assist the design of these systems, namely, NEAT (NeuroEvolution of Augmented Topologies) and HyperNEAT (Hypercube-based NeuroEvolution of Augmented Topologies). These algorithms exploit the evolution of the structure of actuators encoded through neural networks. To evaluate these algorithms, we compare them with a similar approach using the Age Fitness Pareto Optimization (AFPO) algorithm, with a focus on assessing the maximum displacement achieved by the discovered biohybrid morphologies. Additionally, we investigate the effects of optimization against both the volume of these morphologies and the distance they can cover. To further accelerate the computational process, the proposed methodology is implemented in a client-server setting; so, the most demanding calculations can be executed on specialized and efficient hardware. The results indicate that the HyperNEAT-based approach excels in identifying morphologies with minimal volumes that still achieve satisfactory displacement targets.
\end{abstract}


\section{Introduction}

The primary focus of the robotics field is designing and building machines capable of assisting humans across a variety of tasks such as surgery \cite{Heredia2018}, agriculture \cite{Rose2021}, and manufacturing \cite{Evjemo2020}. Soft robots are a sub-category utilizing flexible, deformable materials~\cite{lipson2014challenges,rus2015design,lee2017soft,wang2018toward}, as opposed to traditional rigid robots. These machines mimic the movement and behaviour of living organisms, making them better-suited for various applications, particularly in fields like healthcare, search and rescue, and space exploration. 

Some notable examples of soft robotics based on biological inspiration for their design have been proposed in the past \cite{trivedi2008soft}. For instance, developments of soft robotic suckers inspired by octopus tentacles have been studied~\cite{laschi2012soft}. This highlights the use of flexible materials and smart adhesion. Another study demonstrated a 3D-printed soft robot powered by a combustion reaction~\cite{bartlett20153d}, that showcased the use of innovative materials and manufacturing techniques. Also, a soft robot capable of navigating its environment by growing, in a similar way as a plant was implemented \cite{hawkes2017soft}. There are several original ideas and implementations on plant roots-based inspiration for soft robots \cite{mazzolai2017can,sadeghi2016plant}.

The field of soft robotics continues to advance, with researchers drawing inspiration from biology to create robots that can interact with delicate environments and objects more effectively. As the field matures, it holds great promise for a wide range of applications, and ongoing research continues to push the boundaries of what soft robots can achieve. More recently, robotics not only take inspiration from biology, but even employ biological tissue as part of their functionality \cite{mestre2021biohybrid,morimoto2018biohybrid}.

Given the challenges that arise from utilizing living tissues and the attempt to control them, automated techniques for designing soft robots were studied to unveil the capacity of these novel machines to tackle specific problems. The methodology presented in our current work represents the initial steps towards the development of a biohybrid machine (BHM) actuator that will be utilised for biomedical applications.

Typical robot design is a complex process, since it may imply numerous iterations of testing designs in real life to alleviate the effect of the reality gap in simulations. As a consequence, previous studies required substantial time and material resources to be spent to find the most suitable robot design \cite{Schulz2016}. For soft robots, challenges are even more complex \cite{Hiller2014}. The added complexity primarily arises from the use of flexible materials, characterised by intricate and nonlinear mechanical properties. BHMs integrate biological components, such as cells or tissues, with synthetic elements, a fact that introduces an additional layer of complexity. These BHMs effectively merge the unique capabilities of biological systems with the robust characteristics offered by synthetic materials. However, this amalgamation also adds to the already intricate design landscape by introducing a multitude of parameters that must be carefully considered. 

To tackle these challenges that arise during robot design considerations, an interesting approach was proposed, namely NeuroEvolution (NE). NE involves the evolution of topologies and connection weights of artificial neural networks (ANNs) employing a genetic algorithm (GA). Arguably, the most prominent approach to evolve ANNs is NeuroEvolution of Augmenting Topologies (NEAT) \cite{Stanley2002}, which has been successfully applied in several fields of robotics such as morphology design \cite{Auerbach2011} and gait generation \cite{Reyes2019}. Moreover, NEAT was extended in order to reproduce natural patterns in the learning process. This new approach, called HyperNEAT, employs a type of ANNs called Compositional Pattern-Producing Networks (CPPNs) to generate symmetry and other patterns in the evolved topologies \cite{Stanley2009}. 

The primary aim of our research is to evaluate the applicability of NEAT and HyperNEAT on the BHM design process. We assess the performance of the methodology utilising NEAT and HyperNEAT in comparison to an approach based on Age Fitness Pareto Optimization (AFPO) \cite{Schmidt2010}. The comparison is based on two key metrics of the produced BHM morphologies, namely the maximum displacement achieved within a specified time-frame and the trade-off between the total volume of the morphologies and the displacement attained.

Furthermore, the implementation of the proposed methodology that utilizes NEAT and HyperNEAT has been designed by taking into consideration multiprocessing as a way to accelerate computations. More specific, the implementation is following a client-server model to outsource the more computationally intensive part of the optimization process on distributed and more efficient computing platforms.


\section{Background}\label{sec:background}

NEAT has been widely used in robotics, particularly in locomotion tasks. For instance, NEAT alongside CPPNs were utilised to develop a model that improves the locomotion of snake-like modular robots \cite{Song2023}. That study compared the performance of NEAT against a Multiobjective Genetic Algorithm (MOGA) in three different environments. Results showed that NEAT outperforms MOGA in locomotion control throughout all the predefined environments. Under the same paradigm, {\em Combined-CPPN-NEAT} \cite{Kimura2016} was introduced as a method to design robots capable of performing multiple tasks. This method integrated the topologies of networks that were first evolved to solve single tasks. The method was compared against a MOGA using NEAT in two scenarios: (i) designing morphologies, and (ii) locomotion tasks. Results indicate that Combined-NEAT-CPPN was able to overcome MOGA-NEAT in both scenarios. 

Taking into account works of simulated-only robotics, NEAT has been applied to guide the designing process and the control of robots. For example, the performance of NEAT against the Accuracy-Based Learning Classifier System (XCS) was compared in Robocode, a simulator of tanks \cite{Nidorf2010}. Each tank was driven by a Java agent, which had an extensive range of behaviours and sensory data. Both approaches were compared in scanning and targeting tasks. NEAT demonstrated a suitable performance; however, it required an extensive training period. Also, XCS could learn fast, except in skewed payoff scenarios. Furthermore, under this domain, both approaches were susceptible to over-fitting. In the same direction, NEAT was applied to develop locomotive behaviours of virtual creatures \cite{Tibermacine2014}. This study utilised a physics engine called {\em Open, Dynamic Engine} (ODE) for experimentation. NEAT was compared against a more traditional approach to evolve ANNs \cite{Ruebsamen2002}. Three behaviours were tested: crawling, running, and somersault. Results indicate that the behaviours generated by NEAT tend to be more robust, and they take better advantage of the capacity of the morphologies than those generated by the traditional approach. 

\section{Methods}\label{sec:morph_generator}

The NE algorithms utilized here are population-based optimization methods following GA methodology. Individuals of the populations are networks (ANNs and CPPNs) that encode 3D BHM morphologies. The 3D morphologies are evaluated by a physics engine called {\em Voxelyze} \cite{Hiller2014}. The output of the physics engine used here consists of: (a) the absolute displacement of the morphology from its initial position during the lapse set (10 seconds); and (b) the number of voxels that compose the morphology. Voxels are the elementary building blocks in Voxelyze that can simulate different materials, i.e., active or passive with different parameters (details on the utilization of the simulator are provided in previous works \cite{Kriegman2020,tsompanas2024outline}).



\begin{figure}[tb!]
  \centering
     \includegraphics[width=0.50\linewidth]{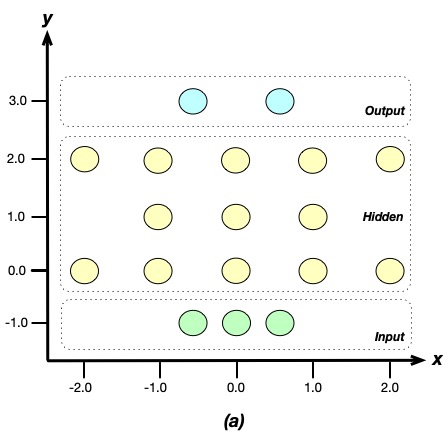}
  \caption{Substrate utilised under HyperNEAT for morphology generator.}
  \label{fig:morph_generator_substrate}
\end{figure}

HyperNEAT evolves CPPNs that in turn design ANNs \cite{Stanley2009}. Thus, based on a three-dimensional layout ($x$, $y$, $z$) 
where BHM morphologies need to be designed, the number of input neurons of the substrate is three. Substrate defines the ANNs that will decode into 3D BHM morphologies. Furthermore, it is necessary to determine the presence or not of a voxel for each position (i.e., a unique ($x$,$y$,$z$) tuple) of the layout and, if present, the type of material the voxel is made of (i.e., active or passive). Thus, the number of output neurons of the substrate is two. Figure~\ref{fig:morph_generator_substrate} depicts the neuron allocation of the substrate utilised during experimentation.

Based on the aforementioned definition, the substrate is queried as follows:

\begin{equation}\label{eq:morph_generator_experimental_setup_substrate}
    SUBSTRATE(x_i,y_i,z_i) =  PV_i, M_i
\end{equation}

\noindent
where $[x_i, y_i, z_i]$ represent the coordinates of the $i$ point in the layout, whereas $PV_i$ represents the presence of a voxel in the $i$ point of the layout, and $M_i$ refers to the type of material the voxel is made of.





It is essential to emphasise that the neuron allocation used for these experiments was chosen after a series of experiments on various numbers of neurons in the $[1,7]$ range per layer and various numbers of layers in the $[1,7]$ range. Regarding the activation function implemented in the substrate, {\em ReLU} was chosen after experimenting with a set of activation functions.


Note here that the NEAT mechanism does not require a substrate to operate. On the contrary, NEAT applies GA operations on the CPPNs that will directly encode the 3D morphologies. 
When applying NEAT, the number of dimensions in the layout where BHM morphologies are designed determines the number of inputs for CPPNs. In this case, the environment is three-dimensional, so, the number of inputs is set to three. Additionally, based on the voxel encoding, the number of output neurons is set to two. Thus, CPPNs are queried as follows: 

\begin{equation}\label{eq:neat_cppn_morph_generator_3_inputs}
    CPPN(x_i, y_i, z_i) = PV_i, M_i
\end{equation}

\noindent
where $[x_i, y_i, z_i]$ refer to the coordinates of the $i$ point in the layout. Regarding $PV_i$, it represents the presence of a voxel in the $i$ point of the layout. Furthermore, $M_i$ represents the type of material the voxel is made of in the $i$ point of the layout. 

In order to analyse the performance of BHM morphologies generated by NEAT and HyperNEAT, two metrics are employed (i) the displacement of the BHM during a simulation of 10 seconds and (ii) the trade-off between displacement and the number of voxels within the simulated BHM (i.e. the total volume of the BHM). The outcomes of both metrics are studied under 500 different controller scenarios (i.e., each active voxel expands at a different phase offset that is generated at random for each scenario), in order to represent the inherent difficulties of controlling biological cells. Furthermore, an approach based on AFPO \cite{Kriegman2020} is used to generate morphologies as a baseline metric to compare the results from NE algorithms. 

\begin{figure}[tb!]
  \centering
     \includegraphics[width=1.0\linewidth]{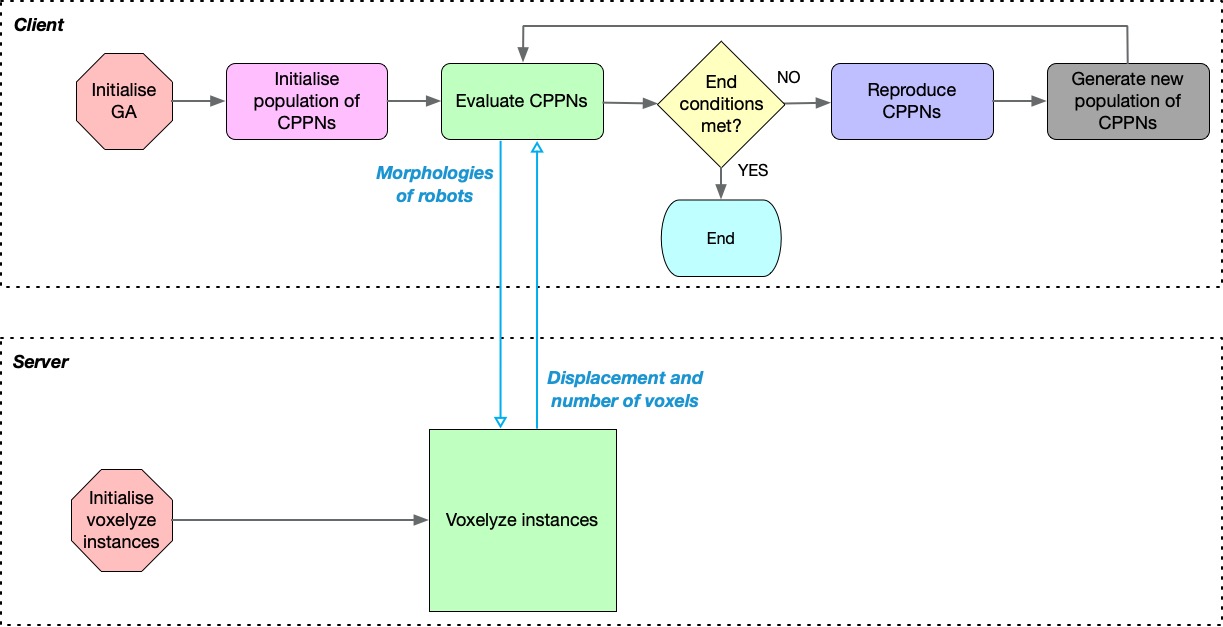}
  \caption{Diagram of the implementation of NEAT and HyperNEAT for producing BHM morphologies.}
  \label{fig:morph_generator_architecture}
\end{figure}

\begin{figure}[tb!]
  \centering
     \includegraphics[width=1.0\linewidth]{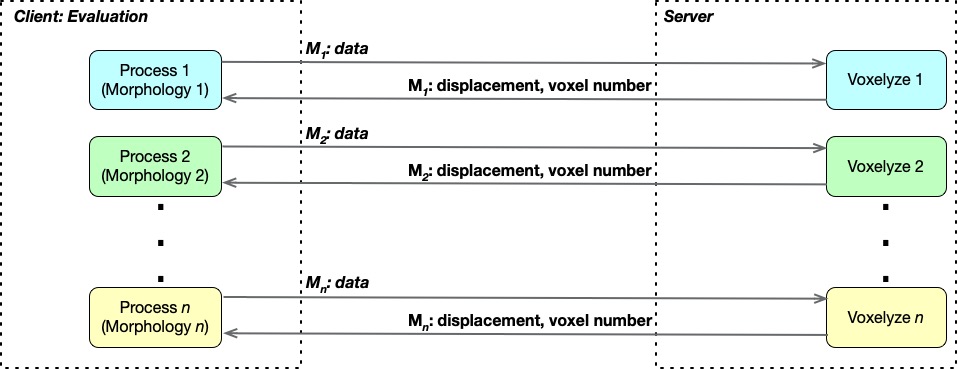}
  \caption{Evaluation of individuals with processes triggered and concurrent interactions with Voxelyze instances.}
  \label{fig:morph_generator_evaluation}
\end{figure}

Due to the considerable computation time required when running the physics engine to evaluate individuals, the code used as a baseline has been written under a multiprocessing paradigm \cite{Kriegman2020}. That is a strategy to reduce the total execution time as much as possible. Our implementation utilising NEAT and HyperNEAT has also been designed considering multiprocessing as a technique to attenuate the time spent for evaluation. In specific, the software has been written under a client-server architecture to take advantage of distributed computing capacities. Figure~\ref{fig:morph_generator_architecture} depicts the general workflow of the implementation. The evaluation of individuals can take place concurrently, which leads to asynchronous execution. Figure~\ref{fig:morph_generator_evaluation} illustrates in more detail the evaluation stage. Thus, the data output of the server-side is utilised to calculate the fitness of morphologies. It is essential to point out that communication is established by HTTP, employing GET and POST methods. JSON is the format for transferring data between the NE GA (executed on the client-side) and the fitness function (executed on the server-side), which includes a reverse proxy acting as a load balancer.

Due to the flexible and scalable nature of the developed software, it is plausible to implement it in several diverse infrastructures. For instance, the server can be deployed in a virtual machine or any cloud computing platform. Regarding the client, it can be deployed regardless of the operative system and the architecture. Moreover, the software takes full advantage of computational power in both ends (i.e., the number of cores in the CPU). Finally, three virtual machines were used here during experimentation, one for each approach. The hardware configured for all virtual machines is described as follows:

\begin{itemize}
    \item {\em Processor}: ARM, 9 cores (18 threads), 3.20GHz.
    \item {\em RAM memory}: 16GB, LPDDR5.
\end{itemize}

\subsection{Experimental set-up}\label{sec:morph_generator_experimental_setup}

For all approaches, the population contains 50 encoded individuals (i.e., CPPNs or ANNs that will translate into 3D actuator morphologies), and each evolutionary run is terminated after 1000 generations. Moreover, the available activation functions are as follows: \textit{sine}, \textit{negative sine}, \textit{absolute}, \textit{negative absolute}, \textit{square}, \textit{negative square}, \textit{square root of absolute}, \textit{negative square root of absolute} and \textit{sigmoid}. The implementation used to generate the AFPO-based morphologies is detailed in \cite{Kriegman2020}, available on GitHub\footnote{\label{code}\url{https://github.com/skriegman/reconfigurable_organisms}} and it was not modified during experimentation. Under the three approaches, the population is initialised following the procedure described in \cite{Kriegman2020} to provide a meaningful comparison of the outputs. Regarding NEAT and HyperNEAT, Table~\ref{tab:morph_generator_parameters} presents the parameters used during the evolution of CPPNs. For each approach, five experimental evolution runs were conducted with different initial populations. 

\begin{table}
\caption{Parameters utilised to evolve CPPNs under NEAT and HyperNEAT in the morphology generator domain.}
\label{tab:morph_generator_parameters}
 \begin{center}
  \begin{tabular}{ c c } 
   \hline
   Parameter & Value \\
   \hline
   compatibility threshold & 3 \\
   compatibility disjoint coefficient & 1.0 \\ 
   compatibility weight coefficient & 0.5 \\
   maximum stagnation & 15 \\
   survival threshold & 0.3 \\ 
   activation function mutate rate & 0.4 \\
   adding/deleting connection rate & 0.3/0.2 \\
   activating/deactivating connection rate & 0.5\\
   adding/deleting node rate & 0.3/0.2 \\
   \hline
  \end{tabular}
 \end{center}
\end{table}

\section{Results}

Since the purpose of the experiments described here is to test the suitability of morphologies generated by AFPO, NEAT and HyperNEAT in terms of locomotion capabilities and robustness, two series of experiments are conducted: (i) finding the most capable morphology in terms of reaching the maximum displacement possible; and (ii) identifying morphologies that reach the maximum displacement possible, in addition to containing the minimum number of voxels. Here the design space is defined by the $x$ and $y$ axis having a range of $[0,8]$ voxels, while the $z$ axis has a range of $[0,7]$ voxels.

\subsection{Maximising displacement }\label{sec:morph_generator_performance_displacement}

One of the crucial features of BHM morphologies is the displacement capability, regardless of the external conditions and under a possibly loosely defined control strategy. Thus, the fittest morphologies generated by AFPO, NEAT and HyperNEAT provided after the 1000 generations, were tested utilising 500 different phase offsets of each of the active voxels in a morphology. These controller phase offset scenarios were {\em a priori} generated (at random) and used consistently for all the morphologies generated by all approaches. It is noteworthy that this experiment does not consider the number of voxels that compose morphologies as a penalty to the fitness. 

\begin{figure*}[tb!]
  \centering
     \includegraphics[width=1.0\linewidth]{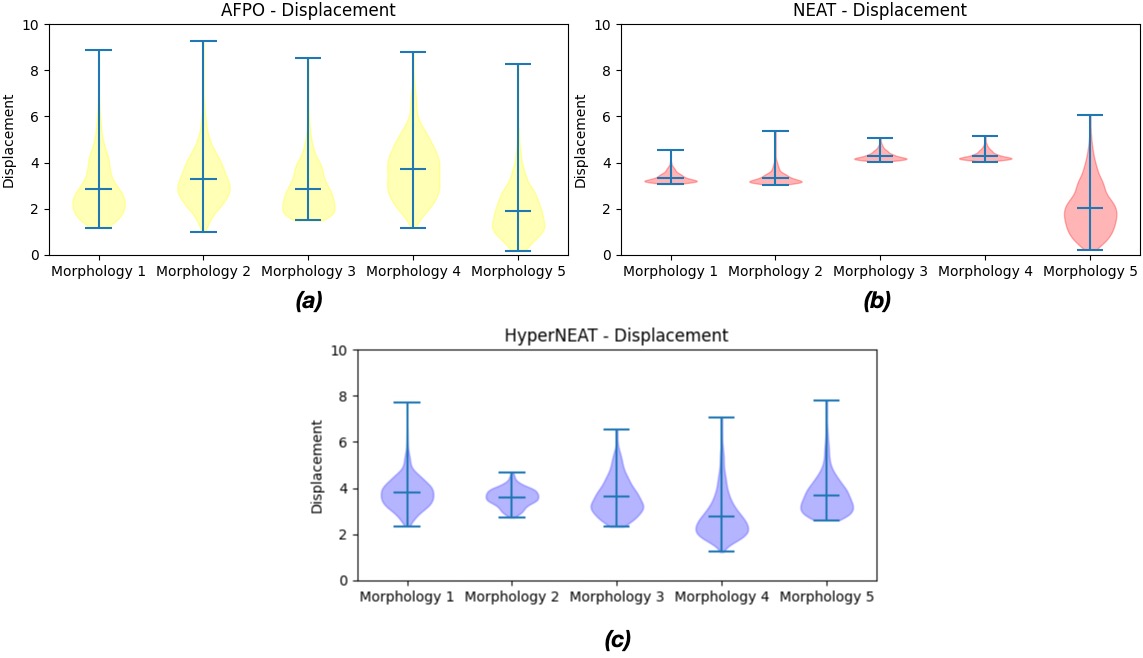}
  \caption{Displacement observed under 500 different phase offset scenarios for the controllers of the fittest morphologies found in five different evolutionary runs under: (a) AFPO-based approach, (b) NEAT-based approach, and (c) HyperNEAT-based approach.}
  \label{fig:morph_generator_performance_displacement}
\end{figure*}

Figure~\ref{fig:morph_generator_performance_displacement} presents violin plots comparing the performance in terms of displacement observed by morphologies generated by AFPO, NEAT, and HyperNEAT. Each violin shows the kernel density estimation of the frequency distribution, maximum, minimum and median of displacement values across the 500 controller runs (i.e., phase offset scenarios). 

Under AFPO (see Fig.~\ref{fig:morph_generator_performance_displacement}-a), all morphologies present significant differences. The produced data are not normally distributed (Shapiro-Wilk test; $p<0.01$). Then, by utilising the Kruskal-Wallis test, it is feasible to state that among the displacements observed, significant differences exist ($p<0.01$). Derived from this result, it is feasible to rank the maximum displacement reached by the morphologies as in: Morphology 4 $>$ Morphology 2 $>$ Morphology 1 $>$ Morphology 3 $>$ Morphology 5 (Dunn’s test: $p<0.01$).

Regarding NEAT (see Fig.~\ref{fig:morph_generator_performance_displacement}-b), similar results are obtained; all morphologies exhibit significant differences. The data obtained are not normally distributed (Shapiro-Wilk test; $p<0.01$). Then, by employing the Kruskal-Wallis test, it can be confirmed that among the displacements obtained, significant differences are emerging ($p<0.01$). Thus, ranking the performance of the morphologies in terms of maximum displacement reached can be performed:  Morphology 4 $>$ Morphology 3 $>$ Morphology 1 $>$ Morphology 2 $>$ Morphology 5 (Dunn’s test: $p<0.01$).

Under HyperNEAT all morphologies exhibit significant differences (see Fig.~\ref{fig:morph_generator_performance_displacement}-c). Again, all collected data are not normally distributed (Shapiro-Wilk test; $p<0.01$). Then, employing the Kruskal-Wallis test, it is possible to confirm significant differences ($p<0.01$). Based on this result, it is plausible to rank the performances (i.e., maximum displacement reached) of the morphologies generated by HyperNEAT: Morphology 1 $>$ Morphology 5 $>$ Morphology 3 $>$ Morphology 2 $>$ Morphology 4 (Dunn’s test: $p<0.01$).

\begin{figure*}[tb!]
  \centering
     \includegraphics[width=0.53\linewidth]{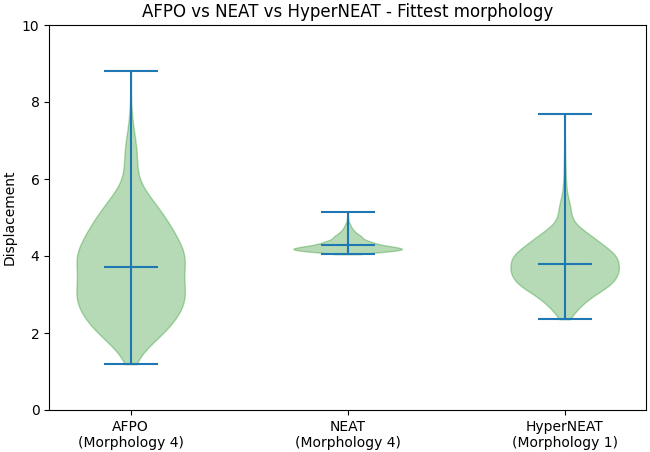}
  \caption{Fittest morphology in terms of reaching maximum displacement under: AFPO, NEAT, and HyperNEAT.}
  \label{fig:morph_generator_benchmark_displacement}
\end{figure*}

Finally, Fig.~\ref{fig:morph_generator_benchmark_displacement} depicts violin plots comparing the one best performing morphology out of the fittest ones generated by AFPO, NEAT, and HyperNEAT in terms of reaching maximum displacement. Each violin shows the kernel density estimation of the frequency distribution, maximum, minimum and median of values across 500 runs (i.e., phase offset controller scenarios). The Kruskal-Wallis test makes it feasible to confirm the existence of significant differences among the fittest morphologies ($p<0.01$). Hence, the performance can be ranked as follows: NEAT $>$ HyperNEAT $>$ AFPO (Dunn’s test: $p<0.01$).

Consequently, the results suggest that, in general, NE-based approaches can produce morphologies that can reach higher displacement than those generated by AFPO, regardless of the phase offset scenario. This can lead to the conclusion that the morphologies produced by NE are more robust. Moreover, when NEAT is compared against HyperNEAT, it can generate morphologies that reach longer displacements and, considering the tight data distribution observed, their performance tend to be more consistent regardless of the controller scenario.

\subsection{Minimising volume and maximising displacement}\label{sec:morph_generator_performance_trade_off}

Although the capability of displacement is crucial, additional aspects may be required to consider a morphology as suitable for production in real-life. Since morphologies represent BHMs that are aiming to be manufactured in minuscule scales (e.g., part of a catheter manoeuvring within human blood vessels), a morphology with fewer voxels (thus, of smaller volume) is considered more suitable and less energy consuming. In order to evaluate the suitability of morphologies, the evaluation will now consider: (a) their capacity on achieving displacement (maximising target); and (b) the amount of voxels comprising them (minimising target). The mechanism to evaluate the displacement of morphologies ($\delta$) is defined as follows:

\begin{equation}\label{eq:biomeld_displacement}
    \delta_{m} = \frac{\Delta_{m}}{\Delta_{max}}
\end{equation}

\noindent
where $\Delta_m$ is the displacement of morphology $m$, which is obtained as an output of the simulation on the physics engine, and $\Delta_{max}$ is the maximum displacement configured for this experiment, which is set to 20. Furthermore, $\delta$ is normalised in the $[0.0,1.0]$ range. The evaluation of the number of voxels of a morphology ($\nu$), is characterised as follows:  

\begin{equation}\label{eq:biomeld_voxels}
    \nu_{m} = 1 - \frac{\Upsilon_{m}}{\Upsilon_{max}}
\end{equation}

\noindent
where $\Upsilon_{m}$ represent the number of voxels contained in morphology $m$ and $\Upsilon_{max}$ is the maximum number of voxels possible, which is given by the dimension of each axis composing the layout where BHMs are designed: $8 \times 8 \times 7 = 448$; consequently, $\Upsilon_{max}=448$. Moreover, $\nu$ is normalised in the $[0.0,1.0]$ range. Once $\delta$ and $\nu$ have been calculated, the fitness of morphology $m$ is obtained as follows:

\begin{equation}\label{eq:biomeld_aptitude}
    fitness_{m} = \frac{1}{2}\delta_m + \frac{1}{2}\nu_m
\end{equation}

\begin{figure*}[tb!]
  \centering
     \includegraphics[width=1.0\linewidth]{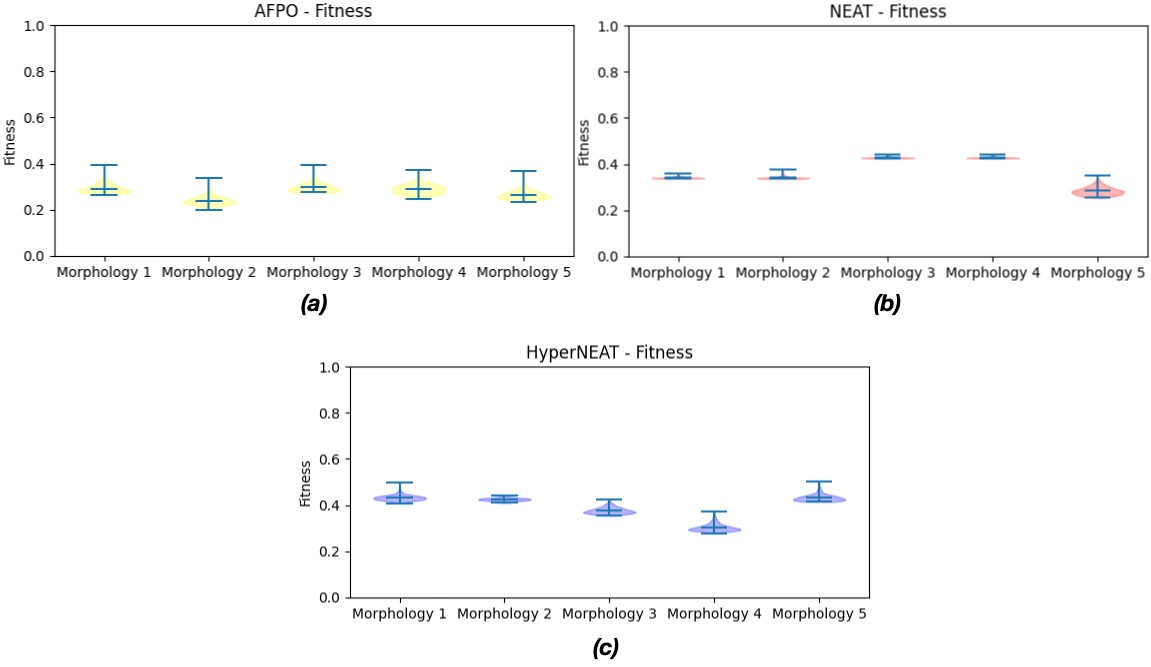}
  \caption{Fitness measured under 500 phase offset scenarios for the controllers of the fittest morphologies found in five different evolutionary runs under: (a) AFPO-based approach, (b) NEAT-based approach, and (c) HyperNEAT-based approach.}
  \label{fig:morph_generator_fitness}
\end{figure*}

\noindent
Figure~\ref{fig:morph_generator_fitness} illustrates violin plots comparing the fitness of morphologies generated by AFPO, NEAT and HyperNEAT under 500 phase offsets (generated {\em a priori} at random) for the new fitness function. Each violin shows the minimum, maximum, median and kernel density estimation of the frequency distribution of values across 500 runs (i.e., phase offset scenarios).

When AFPO is applied (see Fig.~\ref{fig:morph_generator_fitness}-a), all morphologies demonstrate significant differences (Shapiro-Wilk test, Kruskal-Wallis test; $p<0.01$). Consequently, it is feasible to rank the performance of the morphologies created by AFPO: Morphology 3 $>$ Morphology 1 $>$ Morphology 4  $>$ Morphology 5 $>$ Morphology 2 (Dunn’s test: $p<0.01$). Furthermore, under NEAT (see Fig.~\ref{fig:morph_generator_fitness}-b), significant differences can be observed among generated morphologies (Shapiro-Wilk test, Kruskal-Wallis test; $p<0.01$). Thus, the performance of these morphologies can be ranked as: Morphology 4 $>$ Morphology 3 $>$ Morphology 1  $>$ Morphology 2 $>$ Morphology 5 (Dunn’s test: $p<0.01$). Regarding the morphologies generated by HyperNEAT (see Fig.~\ref{fig:morph_generator_fitness}-c), significant differences exist among them (Shapiro-Wilk test, Kruskal-Wallis; $p<0.01$). Hence, a performance rank of the morphologies created by HyperNEAT can be stated: Morphology 5 $>$ Morphology 1 $>$ Morphology 2  $>$ Morphology 3 $>$ Morphology 4 (Dunn’s test: $p<0.01$).

\begin{figure*}[tb!]
  \centering
     \includegraphics[width=0.53\linewidth]{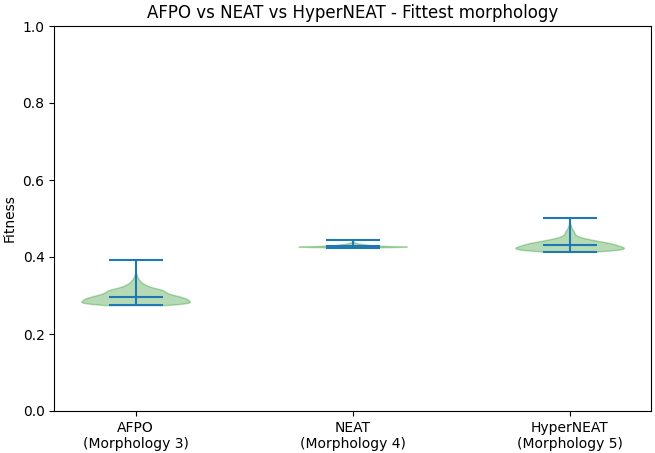}
  \caption{Fittest morphology under: AFPO, NEAT, and HyperNEAT.}
  \label{fig:morph_generator_benchmark_fitness}
\end{figure*}

Figure~\ref{fig:morph_generator_benchmark_fitness} depicts violin plots comparing the performance of the fittest morphology generated by AFPO, NEAT and HyperNEAT. 
The Kruskal-Wallis test allows the suggestion of significant differences among the fittest morphologies ($p<0.01$). Hence, the performance can be ranked as follows: HyperNEAT $>$ NEAT $>$ AFPO (Dunn’s test: $p<0.01$).

The narrow data distribution observed in the NE produced morphologies suggests that their responses are similar, regardless of the controller scenario. This can be interpreted as a capacity to perform similarly, despite the synchronisation of the controller. On the other hand, the morphology created by AFPO has a more wide data distribution, suggesting the morphology has a suitable response in only specific scenarios, where the controller is implemented exactly as designed. Consequently, the robustness of this morphology is not comparable to the ones demonstrated by the morphologies produced by NEAT and HyperNEAT. 

Results indicate that despite the lack of geometrical aspects of the problem domain that could be embodied in the design of the substrate, the morphologies generated by HyperNEAT have a better trade-off between displacement and the number of voxels than those produced by AFPO and NEAT. This can be justified by the fact that HyperNEAT can produce minimal (i.e., lower volume) and robust morphologies with suitable displacement capabilities under several different controller scenarios. In contrast, AFPO and NEAT create morphologies that in some controller scenarios reach a suitable displacement, however, the number of voxels is higher (i.e., higher volume) than that of the HyperNEAT morphologies (see Table~\ref{tab:voxels_afpo_neat_hyperneat}). 

\begin{table}
\caption{Number of voxels, mean displacement and fitness value of fittest morphologies produced by AFPO, NEAT, and HyperNEAT.}
\label{tab:voxels_afpo_neat_hyperneat}
 \begin{center}
  \begin{tabular}{| c | c | c | c |} 
  \hline
   Approach & Number of voxels &  Mean displacement & Fitness value \\
 \hline \hline
   AFPO & 224 & 2.837 & 0.297  \\ \hline
   NEAT & 128 & 4.298 & 0.429 \\ \hline
   HyperNEAT & 116 & 3.686 & 0.432 \\
   \hline
  \end{tabular}
 \end{center}
\end{table}

\begin{figure*}[tb!]
  \centering
     \includegraphics[width=0.71\linewidth]{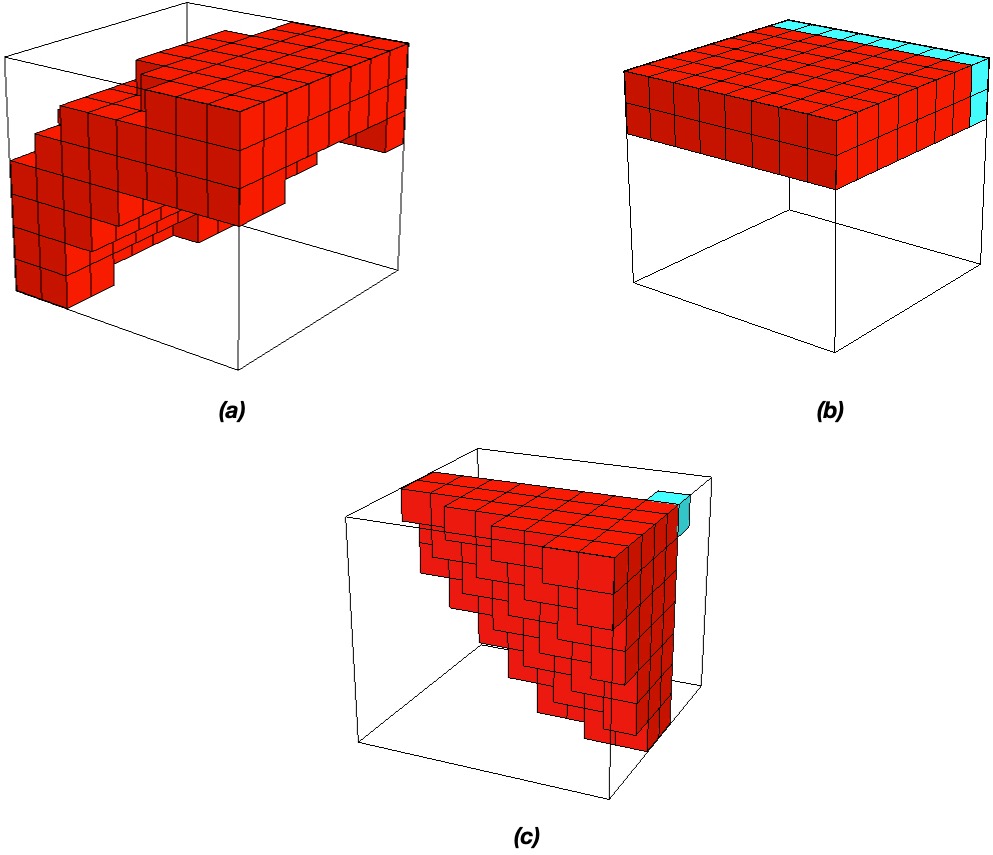}
  \caption{Fittest morphology generated by: (a) AFPO; (b) NEAT; and (c) HyperNEAT.}
  \label{fig:morph_generator_fittest_morphologies}
\end{figure*}

Finally, Fig.~\ref{fig:morph_generator_fittest_morphologies} depicts the fittest morphologies generated by all approaches, with red cubes representing active voxels and blue representing passive ones. Under AFPO (Fig.~\ref{fig:morph_generator_fittest_morphologies}-a), the morphology contains no passive voxels, and its shape is curve-like (i.e., there are no gaps within the morphology). Furthermore, NEAT produced a smaller and thinner morphology (Fig.~\ref{fig:morph_generator_fittest_morphologies}-b), which is a feature that arguably allows it to have a consistent displacement regardless of the phase offset scenario. Despite the morphology being located on the top of the available design area, it is noteworthy that the final displacement is measured from the position that the morphology has after one initial second, allowed for morphologies to settle under gravity (enough to reach the simulated floor) and a further ten simulated seconds (resulting to a simulation of a total of eleven seconds). Regarding HyperNEAT, the morphology generated (Fig.~\ref{fig:morph_generator_fittest_morphologies}-c) is a solid triangular-like prism with one passive voxel at the top. Arguably, the geometry (i.e., the voxel asymmetry towards the top vertex) exhibited by this morphology produces stability during displacement, regardless of the phase offset scenario.

Although all approaches employ the same hardware, the computational time differs. Under AFPO, the mean time spent for each evolutionary run is 14.70 hours, with a standard deviation of 0.15. Furthermore, when NEAT is utilised, the mean time spent for each evolutionary run is 4.74 hours, with a standard deviation of 0.18. Finally, under HyperNEAT, the mean time spent is 6.82 hours with a standard deviation of 0.15 for each evolutionary run. Despite the different execution times (due to increased functionalities of the AFPO baseline code, like saving checkpoints), the robustness of the generated morphologies is the focus of this work. Further investigation is required to compare the efficiency in terms of execution times for the code implementing the alternative methodologies. Arguably, the software architecture implemented for NEAT and HyperNEAT can utilise the hardware more efficiently, since the client-server model is adopted, resulting to significantly reduced computation time during evolution. 


\section{Conclusions}\label{sec:conclusions}

The work presented here explored the suitability of NEAT and HyperNEAT in generating suitable BHM morphologies for maximizing displacement in a physics simulator. The capabilities of NEAT and HyperNEAT were compared against the AFPO approach, a well-known and tested multi-objective optimisation approach.  The fitness functions considered were (i) the maximum displacement and (ii) the trade-off between the number of voxels composing morphologies along with their displacement. Under 500 different scenarios with different predefined phase offsets of the BHM controllers, five morphologies generated by each approach were compared. Results indicate that, generally, the morphologies generated by HyperNEAT require fewer voxels to reach similar or farther displacements than morphologies produced by AFPO and NEAT. Moreover, compact morphologies tend to show a consistent displacement regardless of the phase offset controller scenario; this can be interpreted as those morphologies being more robust than bulky structures. 

From a biomechanical perspective, these results match one of the most common mechanisms of locomotion observed in soft-bodied animals (e.g., earthworms, sea cucumbers, caterpillars, and snails): {\em peristalsis}, which consists of alternating waves of muscle contraction and relaxation move along the body, anchoring the body to the substratum only in widened regions \cite{Yekutieli2009}. 

Future work directions can be outlined based on the results and insights obtained by this research. For instance, one aspect of future work can be adding more periodic activation functions (e.g., \textit{cosine}, \textit{tangent}) to explore additional morphological patterns produced by CPPNs. Furthermore, fine-tuning the GA may improve the performance. For example, increasing the number of individuals and number of generations may be advantageous, whereas minimal initialization of CPPNs have been proved to accelerate the computations and not affect the efficiency of morphologies \cite{tsompanas2024incremental}. Another relevant aspect to consider is the design of the substrate; arguably, a broader exploration of the number of layers, neurons and activation functions may lead to better performance of HyperNEAT. Moreover, an approach that can evolve the geometry of the substrate (i.e., evolve the location of every neuron and the pattern of weights among them) can be implemented to improve the performance of morphologies; this approach is known as {\em ES-HyperNEAT} \cite{Risi2012}.

In conclusion, given the intricacies involved in the design process of BHMs, it is imperative to subject HyperNEAT methodology to further testing on more intricate challenges. This should encompass not only the simulation of the actuator itself, but, also the interactions with the passive components of a desired BHM system, i.e., in applications of medical devices.

\section*{Acknowledgments}

This project has received funding from the European Union’s Horizon Europe research and innovation programme under grant agreement No. 101070328. UWE researchers were funded by the UK Research and Innovation grant No. 10044516.

\section*{Disclosure of Interests.}
The authors have no competing interests to declare that are relevant to the content of this article.

\end{document}